\documentclass[11pt]{article}

\usepackage[preprint]{acl}

\usepackage{times}
\usepackage{latexsym}

\usepackage[T1]{fontenc}

\usepackage[utf8]{inputenc}

\usepackage{microtype}

\usepackage{inconsolata}

\usepackage{graphicx}

\usepackage{booktabs}
\usepackage{multicol}
\usepackage{multirow}
\usepackage{enumitem}
\usepackage[normalem]{ulem}
\usepackage{amsmath}
\usepackage{array}
\usepackage{arydshln}
\newcolumntype{C}[1]{>{\centering\arraybackslash}p{#1}}

\newcommand{\modelname}{\texttt{DIAGPaper}} 
\newcommand{\customizer}{\texttt{Customizer}} 
\newcommand{\rebuttal}{\texttt{Rebuttal}} 
\newcommand{\prioritizer}{\texttt{Prioritizer}} 
\newcommand{\revieweragent}{\texttt{Reviewer}} 
\newcommand{\authoragent}{\texttt{Author}}

\title{\modelname: Diagnosing Valid and Specific Weaknesses in Scientific Papers via Multi-Agent Reasoning}

\author{Zhuoyang Zou, Abolfazl Ansari, Delvin Ce Zhang$^\dagger$, Dongwon Lee, Wenpeng Yin \\ Penn State University; $^\dagger$University of Sheffield \\  \{zhuoyangzou,wenpeng\}@psu.edu}

\begin{document}
\maketitle

\begin{abstract}
\vspace{-1mm}
Paper weakness identification using single-agent or multi-agent LLMs has attracted increasing attention, yet existing approaches exhibit key limitations. Many multi-agent systems simulate human roles at a surface level, missing the underlying criteria that lead experts to assess complementary intellectual aspects of a paper. Moreover, prior methods implicitly assume identified weaknesses are valid, ignoring reviewer bias, misunderstanding, and the critical role of author rebuttals in validating review quality. Finally, most systems output unranked weakness lists, rather than prioritizing the most consequential issues for users.
In this work, we propose \modelname, a novel multi-agent framework that addresses these challenges through three tightly integrated modules. The \customizer{} module simulates human-defined review criteria and instantiates multiple reviewer agents with criterion-specific expertise. The \rebuttal{} module introduces author agents that engage in structured debate with reviewer agents to validate and refine proposed weaknesses. The \prioritizer{} module learns from large-scale human review practices to assess the severity of validated weaknesses and surfaces the top-$K$ severest ones to users.
Experiments on two benchmarks, AAAR and ReviewCritique, demonstrate that \modelname\ substantially outperforms existing methods by producing more valid and more paper-specific weaknesses, while presenting them in a user-oriented, prioritized manner.\footnote{We will release the code and system upon publication.}
\end{abstract}

\vspace{-3mm}
\section{Introduction}
\vspace{-2mm}

Automatic paper weakness identification has attracted increasing attention for two reasons.
First, it poses a fundamental AI challenge, requiring expert-level reasoning over long and complex scientific submissions; second, the rapid growth of paper submissions has placed heavy strain on the peer-review process, creating an urgent need for reliable AI-assisted reviewing.

Despite recent progress, existing systems, such as general-purpose GPT models and review-specific agents \citep{gao2024reviewer2optimizingreviewgeneration,darcy2024margmultiagentreviewgeneration,jin2024agentreview}, still exhibit fundamental limitations.
First, many existing multi-agent systems model review diversity only at a superficial level.
By assigning agents to predefined roles \citep{jin2024agentreview,DBLP08506} or text segments \citep{darcy2024margmultiagentreviewgeneration} rather than explicitly modeling the evaluation criteria that experts apply, these systems fail to capture how human reviewers selectively assess different intellectual aspects of a paper based on their expertise.
Second, prior systems often assume that identified weaknesses are inherently correct \cite{DBLP000100LWGZC0S0L24,chamounautomated}, overlooking the fact that even human reviews can be biased, or mistaken.
In practice, author rebuttals play a critical role in validating and refining review quality, yet this process is underexplored in existing studies.
Third, most systems present weaknesses as unranked lists, failing to prioritize the most consequential issues from a user’s perspective and thus limiting their practical utility.

\begin{table*}[t]
    \centering
    \small
\setlength{\tabcolsep}{1pt}
\begin{tabular}{p{3.3cm}|C{2.0cm}C{2.0cm}C{2.0cm}C{2.0cm}|C{3.0cm}}
         & AgentReview & ReviewAgents & MARS & MARG  & \modelname{} \\
         & {\tiny \citep{jin2024agentreview}} & {\tiny\cite{DBLP08506}} & {\tiny\cite{wang2025marsefficientmultiagentcollaboration}} & {\tiny\citep{darcy2024margmultiagentreviewgeneration}} 
          & (ours)\\\hline
         \multirow{2}{*}{Each agent denotes} & a  human role  & \multirow{2}{*}{a human role} & \multirow{2}{*}{a human role} & \multirow{2}{*}{paper chunk} &  \multirow{2}{*}{a review criterion}\\
         & {\tiny(reviewer, author, AC)} & & & & \\\cdashline{2-6}
         Rebuttal mechanism & review-level & N & N & N &  weakness bullet level\\\cdashline{2-6}
         \multirow{2}{*}{Reviewer agent config.} & fixed, role-based & \multirow{2}{*}{fixed} & fixed, role-based & fixed, chunk-based & dynamic, content-driven\\\cdashline{2-6}
         Severity-oriented output & N & N & N & N & Y\\
    \end{tabular}
    \vspace{-1mm}
    \caption{Novelty of our \modelname{} compared with prior multi-agent paper review systems.}
    \label{tab:noveltysummary}
    \vspace{-3mm}
\end{table*}

To address these challenges, we propose \modelname, a novel multi-agent framework for paper weakness identification that integrates human-inspired reviewing mechanisms through three tightly coupled modules.
The \customizer{} module simulates reviewers’ internal planning processes by identifying paper-specific evaluation criteria and instantiating multiple reviewer agents with criterion-specific expertise, enabling collaborative yet differentiated reviewing behaviors.
The \rebuttal{} module introduces a validity-oriented mechanism by incorporating author agents that engage in structured debate with reviewer agents, allowing initially identified but invalid weaknesses to be challenged, refined, or filtered out.
The \prioritizer{} module focuses on practical deployment by learning from large-scale human review practices to assess the weakness severity and presenting users with the top-$K$ severest issues.

We evaluate \modelname{} on two benchmarks: AAAR~\cite{lou2025aaar10assessingaispotential}, which contains regularly collected human-written reviews for each paper submission, and ReviewCritique~\cite{du2024llmsassistnlpresearchers}, which provides area-chair-labeled valid and invalid review segments.
These benchmarks enable an authentic assessment of a system’s ability to generate genuinely valid, rather than superficially plausible, weaknesses.
Experimental results demonstrate that \modelname{} achieves state-of-the-art performance on both semantic F1 and \textit{Specificity} metrics.
Further analyses show that \modelname{} serves as a general and effective framework that consistently enhances single-agent models by transforming them into multi-agent systems, yielding substantial performance gains.
Notably, \modelname{} can elevate open-source LLMs to performance levels approaching those of closed-source models such as GPT-4o.

In summary, this work makes three contributions:
(i) we introduce \modelname{}, a human-grounded multi-agent framework for paper weakness identification that achieves state-of-the-art performance;
(ii) we show that \modelname{} generalizes across diverse LLM families;
(iii) we provide detailed analyses that elucidate the strengths and limitations of \modelname{} and prior multi-agent review systems.

\vspace{-2mm}
\section{Related Work}
\vspace{-2mm}

\paragraph{Pre-LLM or Non-LLM Automated Review Systems.}
Prior to LLMs, automated paper review was limited and primarily relied on heuristic or template-based methods. \citet{bartoli2016your} generated reviews by reusing and adapting sentences from human-written review corpora, conditioning on target decisions (e.g., accept or reject) and substituting paper-specific terms, resulting in superficially customized but non-evaluative feedback. ReviewRobot~\citep{wangeviewrobot} employed information extraction and knowledge graphs to identify structured elements from papers and related work, producing category-level comments with explicit evidence. Despite offering limited assistance, these systems lacked deep contextual understanding and could not support constructive evaluation.

\vspace{-2mm}
\paragraph{Single-Agent LLM-Based Review Systems.}
The advent of LLM enabled more holistic automated review attempts. Early GPT-prompting systems \cite{liang2023largelanguagemodelsprovide,zhou2024llm} demonstrated partial success but exposed fundamental limitations in reliability and global judgment. Recent work improved technical coverage while retaining single-agent constraints. REVIEWER2~\citep{gao2024reviewer2optimizingreviewgeneration} introduced question-guided prompts and multi-stage generation to better handle long documents, and both OpenReviewer~\citep{kuznetsov2024openreviewer} and DeepReviewer \cite{DBLPZhuWY025} fine-tuned LLMs on large-scale human reviews.

\vspace{-2mm}
\paragraph{Multi-Agent Systems for Paper Review.}
Multi-agent architectures have been proposed to address single-agent limitations by distributing analysis across specialized agents. AgentReview~\citep{jin2024agentreview} simulates reviewers, authors, and area chairs across multiple review phases to study bias and decision variability. ReviewAgent \cite{DBLP08506} decomposes the review process into a fixed set of trained reviewer and area-chair agents that follow a predefined, human-like review workflow to generate full review reports. MARS \cite{wang2025marsefficientmultiagentcollaboration} assigns multiple independent reviewer agents to evaluate a candidate answer and a meta-reviewer agent to aggregate their feedback.
MARG~\citep{darcy2024margmultiagentreviewgeneration} decomposes papers into chunks processed by different agents.

Despite these advances, existing systems share two key limitations.
First, their agent configurations are largely fixed and do not adapt to paper-specific content or contribution types.
Second, most prior work emphasizes human-like review generation, while lacking a systematic mechanism to validate the correctness of identified critiques.
In contrast, \modelname{} focuses explicitly on paper weakness identification, introducing a validity-oriented, rebuttal-based mechanism to filter incorrect or weakly grounded critiques.
A detailed comparison highlighting the novelty of \modelname{} relative to prior systems is provided in Table~\ref{tab:noveltysummary}.

\begin{figure}[t]
\centering
\includegraphics[width=0.99\linewidth]{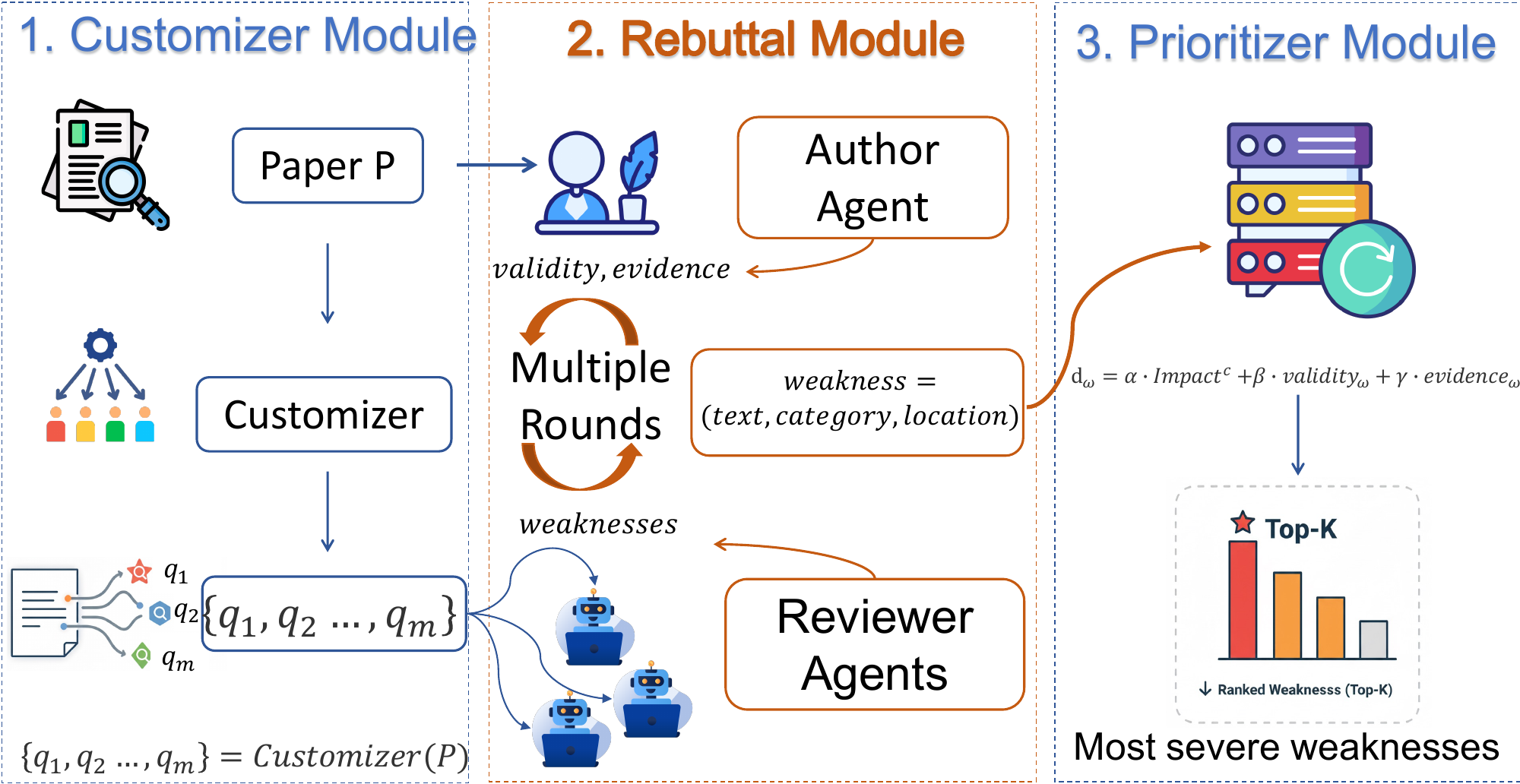} 
\caption{Overview of our \modelname~framework.}
\label{fig:overallsystem}
\vspace{-2mm}
\end{figure}

\section{Method: \modelname}
\vspace{-2mm}

Our system, \modelname{} (Figure~\ref{fig:overallsystem}), comprises three modules for \emph{validity-controlled weakness identification} by simulating key human review behaviors.
The \textbf{\customizer{}} module explicitly models \emph{expertise separation}, decomposing review into fine-grained evaluation aspects (e.g., dataset representativeness, experiment completeness, etc.) to enable principled multi-agent orchestration.
The \textbf{\rebuttal{}} module introduces reviewer--author interactions that challenge and validate proposed weaknesses, mirroring how critiques are regulated in real peer review.
Finally, the \textbf{\prioritizer{}} module ranks validated weaknesses and outputs the top-$K$ most consequential issues, supporting realistic use cases where the severest feedback must be surfaced first.

\vspace{-2mm}
\subsection{\customizer{} Module: Criteria-Oriented Reviewer Decomposition}

\paragraph{Rationale.} Prior multi-agent systems for scientific document analysis typically either configure agents to simulate high-level human roles such as reviewers or area chairs \cite{jin2024agentreview,DBLP08506,wang2025marsefficientmultiagentcollaboration} or  assign each chunk to an individual agent \cite{darcy2024margmultiagentreviewgeneration}. While these approaches yield preliminary performance gains, they model paper review largely \textit{at a surface level} by introducing multiple agents with different apparent roles. In contrast, human review diversity stems from reviewers’ heterogeneous backgrounds and expertise, which lead them to apply different criteria when evaluating the same submission. By focusing on roles rather than explicitly modeling and controlling such criteria, prior systems fail to capture how expert reviewers \textit{selectively assess different intellectual aspects} of a paper.

This  \customizer{} module is designed to emulate how human reviewers internally plan their evaluation when encountering a new submission. Based on the paper’s concrete content and claimed contributions, reviewers implicitly determine which dimensions require closer scrutiny. For instance, when a paper’s primary contribution is a new dataset, reviewers often focus on aspects such as scalability of the data collection process, quality control, and the novelty and necessity of the dataset relative to existing resources.

\vspace{-2mm}
\paragraph{Approach.} To operationalize this process, we first engaged three AI professors  to crowdsource a set of commonly used review dimensions and jointly distill a category for each. Through iterative discussion, the experts refined the set until reaching consensus on its completeness. This effort resulted in 20 review dimensions, each capturing a core criterion for evaluating the quality of a paper submission. Two examples are ``\textit{[Method--Clarity] Any unclear or confusing part in the approach description?}'' and ``\textit{[In-depth Analysis] Do the experimental analyses provide sufficient insight and explanation, or merely describe superficial phenomena?}''. The full list of review dimensions is provided in Table~\ref{append:review_criteria} in Appendix~\ref{sec:append:dimensions}.
These dimensions are used to instantiate a corresponding set of \revieweragent{} agents—each responsible for one dimension—in the subsequent \rebuttal{} stage.

To enable a fully automated system, \modelname{} further introduces a dedicated \customizer{} agent that dynamically generates paper-specific review dimensions without relying on manually predefined categories. Formally, the module is defined as
\vspace{-2.5mm}
\begin{equation}
    \{q_1, q_2, \ldots, q_m\} = \customizer(P),
    \vspace{-2.5mm}
\end{equation}
where $P$ denotes the paper submission and each $q_i \in D$ represents a question-style review dimension. In our experiments ($\mathcal{Q}_3$ in Section \ref{sec:analysis}), we analyze and compare \customizer{}-generated dimensions ($D_c$) with human-curated dimensions ($D_h$).

\vspace{-2mm}
\subsection{\rebuttal{} Module: Adversarial Interaction Between \revieweragent{} and \authoragent{}}\label{sec:rebuttal}

\paragraph{Rationale.}
A desirable rebuttal mechanism should go beyond providing auxiliary information for meta-review, but instead serve as a \emph{validity-regulating process for reviewer critiques}. Here, validity refers to whether an identified weakness is \emph{well-grounded in the paper content and supported by concrete, localized evidence}, rather than verifying the factual correctness of the paper’s scientific claims.
In practice, authors contest individual weakness points rather than entire reviews, and effective rebuttals operate at the granularity of specific critique bullets rather than review-level narratives.
Moreover, rebuttal typically requires grounding arguments in paper-specific evidence and proceeds through iterative, multi-round refinement rather than a single exchange.
These properties motivate a new \rebuttal{} design, in contrast to the only related work AgentReview~\cite{jin2024agentreview}, that is weakness-level, evidence-driven, and multi-round, enabling systematic filtering and refinement of mis-grounded critiques.

\vspace{-3mm}
\paragraph{Approach.}
The \rebuttal{} module improves the reliability of identified weaknesses by explicitly modeling an adversarial reviewer--author interaction, in which an \authoragent{} challenges each claim proposed by \revieweragent{} agents.

\textbullet\enspace\textbf{\revieweragent{} Agents.}
For each review dimension $q_i$ produced by the \customizer{} module, a specialized \revieweragent{} agent identifies potential weaknesses specific to that dimension. Each identified weakness is represented as
\vspace{-2mm}
\begin{equation}
    \textit{weakness} = (\textit{text}, \textit{category}, \textit{location}),
    \vspace{-3mm}
\end{equation}
where \textit{text} describes the critique, \textit{category} denotes the type of issue (e.g., methodological novelty or experimental rigor), and \textit{location} indicates the relevant part of the paper.

\textbullet\enspace\textbf{\authoragent{} Agent.}
Each proposed $weakness$ is then evaluated by an \authoragent{}, which examines the full paper to assess whether the critique is appropriately supported by the submission itself.
This design departs from conventional review aggregation by subjecting each weakness to adversarial, paper-grounded scrutiny rather than accepting reviewer claims at face value.

Given a $weakness$ and the paper $P$, the \authoragent{} produces
\vspace{-3mm}
\begin{equation}
    (\textit{valid}, \textit{evid}) = \authoragent(\textit{weakness}, P),
\vspace{-3mm}\end{equation}
where \textit{valid} is a three-class label (\emph{fully valid}, \emph{partially valid}, \emph{invalid}) indicating the degree to which the critique is justified, and \textit{evid} (evidence) reflects the strength of \emph{paper-grounded support} (\emph{substantial}, \emph{moderate}, \emph{weak/absent}).

The \revieweragent{} and \authoragent{} engage in bounded, multi-round interaction until one of the following conditions is met:
(i) the \authoragent{} provides a fully valid response with substantial paper-grounded evidence that the \revieweragent{} cannot counter;
(ii) both agents agree on partial validity with moderate support; or
(iii) after three rounds\footnote{We cap interactions at three rounds to balance deliberation depth and efficiency, as most disagreements converge within two to three rounds.}, the combined validity and evidence score falls below a threshold ($<\epsilon$).
Weaknesses with insufficient combined support are filtered out.
Empirically, this process removes $\sim$ 40\%--60\% of initially identified weaknesses.

\vspace{-2mm}
\subsection{\prioritizer{} Module: Outputting Top-$K$ Severest Weaknesses}

\paragraph{Rationale.} For weaknesses that pass the \rebuttal{} module, simply reporting them without differentiation is suboptimal from a user perspective. Authors typically prefer to see the most consequential issues first. Accordingly, the primary challenge of the \prioritizer{} module is to quantify the potential $severity$ of each weakness on a paper’s evaluation outcome.

This design is motivated by established peer-review practice. After individual reviews and author rebuttals, final acceptance decisions are largely determined by area chairs or meta-reviewers, who synthesize the paper, the reviews, and the rebuttal to identify the most decisive concerns. Intuitively, weakness categories that are more frequently emphasized in meta-reviews are more likely to influence final decisions.

\vspace{-2mm}
\paragraph{Approach.} Concretely, we analyze how different weakness categories propagate through real conference review processes and define a category-level impact score as $\mathrm{Imp}^{c} = f^{c}_{\mathrm{meta}}/f^{c}_{\mathrm{ind}}$,
where $f^{c}_{\mathrm{meta}}$ and $f^{c}_{\mathrm{ind}}$ denote the frequencies with which weakness category $c$ appears in meta-reviews and individual reviews, respectively. These impact values are learned from  analyzing 15,000+ papers from ICLR, NeurIPS, and ICML
(2020-2024).

Each surviving weakness $w$, associated with category $c$, is then ranked by a severity score $s_w$:
\vspace{-2mm}
\begin{equation}\label{eq:damagescore}
    s_w = \alpha \cdot \mathrm{Imp}^{c} + \beta \cdot \mathrm{valid}_w + (1-\alpha-\beta) \cdot \mathrm{evid}_w,
\vspace{-2mm}\end{equation}
where $\alpha$ and $\beta$ are hyperparameters. Finally our system outputs top-$K$ weaknesses by $s_w$ scores.
This formulation integrates empirical patterns from historical reviews with the adversarial assessments produced by the \rebuttal{} module, yielding rankings grounded in both real reviewer behavior and the intrinsic characteristics of each weakness.

\section{Experiments}

\paragraph{Datasets.}
Our evaluation explicitly accounts for a critical real-world challenge: human-written review weaknesses are often unreliable due to reviewer bias, missing evidence, limited expertise, or superficial judgments. To address this issue, we adopt two complementary datasets with distinct evaluation objectives.

\textbullet\enspace\textbf{AAAR} \citep{lou2025aaar10assessingaispotential}.
A large-scale collection of peer reviews spanning multiple AI domains, including machine learning, computer vision, NLP, robotics, etc. It contains human-written reviews from major conferences, together with both individual reviewer assessments and meta-reviews. The dataset includes 993 papers and 3,759 reviews, with an average of 18,211 segments per review. We use AAAR to evaluate the extent to which our system’s identified weaknesses align with observed human review patterns.

\textbullet\enspace\textbf{ReviewCritique} \citep{du2024llmsassistnlpresearchers}.
A curated benchmark of 100 papers in which area chairs annotate each individual \textit{review segment} as \emph{valid} (ReviewCritique(Valid)) or \emph{invalid} (ReviewCritique(Invalid)).
The valid subset contains 1,374 review segments, while the invalid subset contains 364 segments.
This dataset enables a validity-oriented evaluation, allowing us to measure whether our system produces a higher proportion of valid weaknesses and fewer invalid ones compared to baselines, rather than merely matching potentially flawed  reviews.

\vspace{-2mm}
\paragraph{Evaluation Metrics.}
We adopt the official evaluation metrics associated with each dataset. Specifically, we report semantic-oriented \textit{precision}, \textit{recall}, and \textit{F1}, which measure semantic alignment between system-generated weaknesses and reviewer-written weaknesses. In addition, we report \textit{specificity} developed in \citep{du2024llmsassistnlpresearchers}; it is computed using an inverse TF--IDF–based formulation, quantifying how unique and paper-specific a weakness is by comparing it against weaknesses identified across all reviewed papers.

Importantly, since the \textit{ReviewCritique(Invalid)} split consists of human-written but invalid weakness statements, \emph{lower} semantic overlap (e.g., lower $F1_{\text{invalid}}$) indicates better performance. 
However, this naive metric is brittle: a poor-quality LLM may generate incoherent or previously unseen invalid statements that do not semantically match any annotated invalid weaknesses, thereby achieving an artificially low $F1_{\text{invalid}}$ despite producing outputs that are not review-like or meaningful.
To address this failure mode, we observe that such systems, while avoiding overlap with known invalid weaknesses, also tend to perform poorly on \textit{ReviewCritique(Valid)}, as their outputs lack coherence and grounding. This deficiency is reflected in low precision and recall on valid weaknesses. We therefore introduce a normalized diagnostic metric:
\begin{equation}
    \widetilde{F1}_{\text{inv}} = \frac{F1_{\text{invalid}}}{\text{P}_{\text{valid}} \cdot \text{R}_{\text{valid}}}.
\end{equation}
Here, $F1_{\text{invalid}}$ measures overlap with \emph{known} invalid weaknesses, while $\text{P}_{\text{valid}} \cdot \text{R}_{\text{valid}}$ serves as a proxy for a system’s ability to generate coherent, review-like, and paper-grounded critiques. As a result, systems that achieve low $F1_{\text{invalid}}$ merely by producing incoherent or spurious outputs are penalized due to their low performance on valid weaknesses, whereas systems that consistently generate meaningful critiques are rewarded. We provide a concrete illustrative example and detailed analysis in Section~\ref{sec:mainresults}.

\vspace{-2mm}
\paragraph{Baselines.}
We consider two categories of baselines. i) \textbf{General-purpose LLMs.}
This category includes three open-source LLMs—Mistral-7B \cite{jiang2023mistral7b}, Llama~3.1-70B \cite{grattafiori2024llama3herdmodels}, and Qwen2.5-72B \cite{qwen2025qwen25technicalreport}—and one closed-source model, GPT-4o. These models represent four major LLM families, with GPT-4o widely serving as a  state-of-the-art baseline. ii) \textbf{Review-specific agents}, including single-agent \textit{Reviewer2} \citep{gao2024reviewer2optimizingreviewgeneration}, and two multi-agent systems \textit{AgentReview} \citep{jin2024agentreview} and \textit{MARG} \citep{darcy2024margmultiagentreviewgeneration}.

\vspace{-2mm}
\paragraph{Setup.} The threshold $\tau=0.4$ used in the rebuttal (Section \ref{sec:rebuttal}) is empirically calibrated using large-scale review–revision pairs, leveraging submission editing behavior as a proxy signal for author acceptance of reviewer-identified weaknesses. Parameters in Equation \ref{eq:damagescore} are set to: $\alpha=0.5$, $\beta=0.3$.

\subsection{Main Results}\label{sec:mainresults}

\begin{table*}[t]
\centering
\resizebox{\textwidth}{!}{
\begin{tabular}{l@{\hskip 8pt}cccc@{\hskip 12pt}cccc@{\hskip 12pt}ccc}
\toprule
\multirow{2}{*}{\textbf{Method}} & \multicolumn{4}{c}{\textbf{AAAR}} & \multicolumn{4}{c}{\textbf{ReviewC.(Valid)}} & \multicolumn{3}{c}{\textbf{ReviewC.(Invalid)}} \\
\cmidrule(lr){2-5} \cmidrule(lr){6-9} \cmidrule(lr){10-12}
& P ($\uparrow$) & R ($\uparrow$) & F1 ($\uparrow$) & Sp. ($\uparrow$) & P ($\uparrow$) & R ($\uparrow$) & F1 ($\uparrow$) & Sp. ($\uparrow$) &  F1 & $\widetilde{F1}_{\text{inv}}$ (\textcolor{red}{$\downarrow$}) & Sp. ($\uparrow$) \\
\midrule
\multicolumn{12}{l}{\textit{General-purpose LLMs}} \\
Llama 3.1-70B & 41.02 & 44.89 & 42.78 & 8.32 & 35.66 & 32.96 & 33.00 & 5.02 & \sout{\textnormal{29.52}}& 25.10 & 5.21 \\
Mistral-7B & 40.15 & 44.21 & 42.03 & 9.97 & 41.37 & 46.96 & 43.55 & 9.26 & \sout{\textnormal{38.91}} & 20.01 & 8.64 \\
Qwen2.5-72B & 40.89 & 44.89 & 42.74 & 9.21 & 36.14 & 32.07 & 32.85 & 4.15 &  \sout{\textnormal{22.85}}& 19.71 & 6.65 \\
GPT-4o & 45.71 & 50.35 & 47.43 & 10.52 & 44.08 & 49.74 & 46.24 & 9.39 & \sout{\textnormal{41.11}} & 18.80 & 8.81 \\
\midrule
\multicolumn{12}{l}{\textit{Review-specific agents}} \\
AgentReview & 38.47 & 40.06 & 38.47& 5.91 & 43.99 & 39.32 & 41.15 & 5.16 & \sout{\textnormal{37.65}}& 21.77 & 4.79 \\
MARG & 44.08 & 49.74 & 46.24 & 10.04 & 44.43 & 47.52 & 45.11 & 8.19 & \sout{\textnormal{42.73}}& 20.20 & 8.07 \\
Reviewer2 & 38.37 & 52.21 & 44.12 & 12.15 & 43.41 & 46.06 & 44.07 & 9.72 & \sout{\textnormal{41.30}}& 20.70 & 9.45 \\
\midrule
\textbf{\modelname} & \textbf{49.39} & \textbf{54.87} & \textbf{51.89} & \textbf{13.50} & \textbf{51.21} & \textbf{52.32} & \textbf{50.23} & \textbf{10.25} & --& \textbf{14.92} & \textbf{9.89} \\
\bottomrule
\end{tabular}
}
\caption{\modelname{} vs. baselines on AAAR and ReviewCritique datasets. $\uparrow$: the higher the better; \textcolor{red}{$\downarrow$}: the lower the better. Please note F1 on ReviewCritique(Invalid) is misleading; we add it just for reference. }
\label{tab:main_results}
\vspace{-2mm}
\end{table*}

\begin{table}[t]
\centering
\begin{tabular}{lcccccc}
\toprule
\textbf{Models} & \textbf{Valid}  &\textbf{Spec.} & \textbf{Realism} \\
\midrule
MARG & 62.5& 1.98& 67.8\\
AutoGen-Core    &62.5  &  2.15 &57.1\\
GPT-4o          & 71.4  &2.01 &\textbf{85.7}\\
Reviewer2       & \textbf{88.9}  &2.32 &77.8\\
\hline
\textbf{\modelname{}}    &\textbf{88.9}  &\textbf{4.49} &67.8\\
\bottomrule
\end{tabular}
\caption{Human evaluation for unseen weaknesses.}
\label{tab:human_eval_results}
\vspace{-2mm}
\end{table}

Table~\ref{tab:main_results} summarizes the main results comparing \modelname{} with aforementioned baselines. We highlight three key observations.

First, on \textit{AAAR}, \textit{ReviewCritique(Valid)}, and \textit{ReviewCritique(Invalid)}, \modelname{} achieves state-of-the-art performance across semantic recall, precision, F1, and \textit{Specificity}, demonstrating both strong alignment with valid human reviews and the ability to produce paper-specific weaknesses.

Second, F1 results on \textit{ReviewCritique(Invalid)} exhibit a qualitatively different pattern. Since this split consists of human-written but invalid weakness statements, matching them indicates negative signals. We observe that top-performing systems—GPT-4o and  multi-agent approaches (Reviewer2, AgentReview, MARG, and \modelname{})—exhibit low overlap with these invalid reviews for a desirable reason: they tend to generate valid, well-grounded weaknesses that do not align with invalid human reviews. In contrast, single-agent open-source models (Llama~3.1, Mistral, and Qwen) also show low F1, but primarily because they mostly produce noisy and unseen outputs that fail to meaningfully correspond to either valid or invalid weaknesses. For example, Mistral generated ``\textit{The paper claims COCO-LM improves accuracy on GLUE and SQuAD but offers no significant contributions in methodology or theory},'' which is factually incorrect, matching neither valid nor invalid human reviews. 

Finally, under our diagnostic metric $\widetilde{F1}_{\text{inv}}$, \modelname{} demonstrates the strongest rejection of invalid weaknesses while maintaining the highest \textit{Specificity} score (9.89). 
This result validates the effectiveness of $\widetilde{F1}_{\text{inv}}$ in distinguishing genuinely high-quality review behavior from artifactually low overlap caused by noisy or non-review-like outputs, and confirms the reliable state-of-the-art performance of \modelname{}.

\vspace{-2mm}
\subsection{Analyses}\label{sec:analysis}

Beyond the main comparison in Table~\ref{tab:main_results}, this subsection further investigates four research questions.

\vspace{-2mm}
\paragraph{$\mathcal{Q}_1$: Does multi-agentization of a single LLM improve performance (i.e., cross-LLM generalization)?}
\modelname{} is a framework that can be instantiated with different underlying LLMs. This question examines whether wrapping a single-agent LLM into the \modelname{} multi-agent architecture consistently improves its performance across model families.

\begin{figure}[t]
\centering
\includegraphics[width=0.47\textwidth]{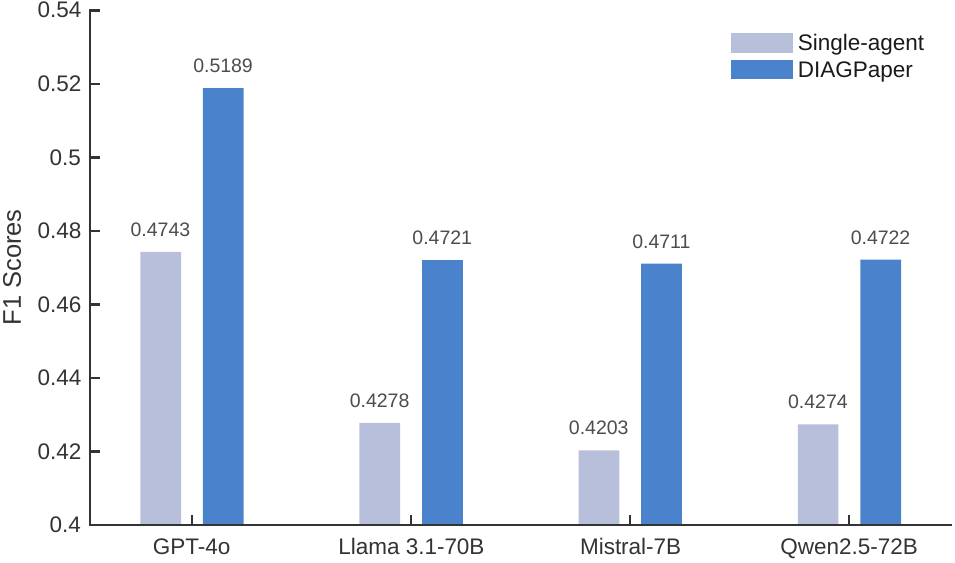} 
\caption{Improvement when converting a single-agent into multi-agent under our \modelname~framework.}
\label{fig:singlevsmulti}
\vspace{-2mm}
\end{figure}

Figure~\ref{fig:singlevsmulti} compares single-agent and multi-agent instantiations under the \modelname{} framework. Across all tested LLMs—including GPT-4o, Llama, Mistral, and Qwen—multi-agentization yields consistent performance gains. \textit{Notably, multi-agent versions built on open-source LLMs (Llama, Mistral, and Qwen) achieve scores above 47.21\%, approaching or matching the performance of the closed-source single-agent GPT-4o (47.43\%)}. These results highlight the framework’s strong cross-LLM generalization and show that users can obtain high-quality review generation without paying costly closed-source LLM API fees, which in turn makes large-scale review of submissions feasible, such as potential deployment in community platforms like OpenReview.

\begin{table*}[t]
\centering
\small
\begin{tabular}{@{}p{0.1\textwidth}p{0.3\textwidth}p{0.40\textwidth}}
\toprule
\textbf{Category} & \textbf{Expert-written} & \textbf{Agent-generated} \\
\midrule
\multicolumn{3}{l}{\textit{(i) \textcolor{blue}{Both Human and Agent Addressed, but Agent Provided Better Analysis ($\sim$ 35\% across the dataset)}}} \\
\midrule
Method-Validity & 
Any methodological flaws or inconsistencies in the proposed approach that could invalidate the results?
 & 
 How does auxiliary model training affect main model convergence? Are gradient flows optimized?\\
\midrule
Writing &
What severe writing issues are making this paper difficult to understand? &
 (1) Are tables/figures clearly formatted? (2) Are technical terms defined? (3) Is content accessible?\\
\midrule
\multicolumn{3}{l}{\textit{(ii) \textcolor{blue}{Human Had No Specific Criteria, Agent Identified Critical Gaps ($\sim$ 30\% across the dataset)}}} \\
\midrule
Statistical Rigor & ---
 &Are results reported with error variance? Are improvements statistically significant?
\\
\bottomrule
\end{tabular}
\caption{Cases when Reviewer agents complement expert-written review dimensions.
}
\label{tab:category_weaknesses}
\vspace{-2mm}
\end{table*}

\vspace{-2mm}
\paragraph{$\mathcal{Q}_2$: Human Evaluation of Low-Precision Weaknesses.}
Our automatic metrics (semantic recall, precision, and F1) measure overlap with human-written weaknesses. However, due to inherent limitations of paper reviewing, such as bias, incomplete coverage, or limited expertise, a semantic mismatch does not necessarily imply that a newly generated weakness is invalid. To assess the quality of such unmatched weaknesses, we conduct a targeted human evaluation.

We hire three senior Ph.D. students to manually annotate 50 low-precision weakness instances produced by \modelname{} and competitive baselines. We exclude clearly underperforming open-source single-agent models (Llama, Mistral, and Qwen) from this analysis. Each weakness' final label is determined via majority vote. The evaluation aims to determine which system is more likely to surface \emph{novel yet reasonable} weaknesses not previously identified by human reviewers.

Annotators are asked to label each weakness along two dimensions: (i) \textit{Validity}—whether the weakness correctly identifies a real issue in the submission; and (ii) \textit{Realism}—whether addressing the weakness is feasible within typical constraints of time and resources. In addition, we report \textit{Specificity} to assess whether unmatched weaknesses are paper-specific rather than generic or random.

Table~\ref{tab:human_eval_results} reports the human evaluation results. Despite focusing on weaknesses not identified by original reviewers, \modelname{} achieves state-of-the-art performance on \textit{Validity} and \textit{Specificity}. However, its \textit{Realism} score is lower than some baselines. Further analysis reveals that this gap primarily arises from \modelname{}’s tendency to \textit{generate overly strict} but factually grounded critiques. We explain this further with the error analysis below.

\vspace{-2mm}
\paragraph{Error Pattern Analysis.}
We further analyze common error patterns across systems and identify the following trends:

\textbullet\enspace\textbf{\modelname{} acts as a stricter reviewer.}
    The dominant failure mode for \modelname{} (71.4\%) involves weaknesses that are factually plausible but judged invalid by annotators because they impose unrealistic expectations beyond the paper’s intended scope. Typical examples include demands for large-scale experiments, extensive additional baselines, or insisting that certain topics be addressed in specific sections. For instance, while a question such as ``Does the paper discuss multimodal extensions?'' is reasonable, \modelname{} may escalate it into an overly strict requirement (e.g., insisting such discussion is mandatory because it appears in future work), which annotators often deem unrealistic. 

\textbullet\enspace\textbf{Incorrect section referencing in \modelname{}.}
    This issue arises from incorrect section attribution. After retrieval and refinement, a claim may appear in multiple parts of the paper; however, \modelname{} occasionally anchors the weakness to an incorrect retrieved passage, leading to mismatches between the critique and the cited context.

\textbullet\enspace\textbf{Baselines flag already acknowledged limitations by 7.1\%; rarely observed in \modelname{}).}
    Some baseline systems identify limitations that are explicitly acknowledged by the authors (e.g., listed as limitations or future work). This pattern is rarely observed in \modelname{}.

\vspace{-2mm}
\paragraph{$\mathcal{Q}_3$: Why does \modelname{} succeed, and does its internal behavior match our design intent?}
To answer this question, we examine the internal behaviors of the three core modules in \modelname{}: \customizer{}, \rebuttal{}, and \prioritizer{}.

\textbf{First, the \customizer{} module.}
We evaluate whether the \customizer{} agent provides advantages over fixed human-written criteria by comparing it with the 20 predefined dimensions (Table~\ref{append:review_criteria}).
Table~\ref{tab:category_weaknesses} presents a case study on the COCO-LM paper~\citep{meng2021cocolmcorrectingcontrastingtext}, while the reported proportions are computed over the full dataset.
Overall, \customizer{} (i) identifies more paper-specific evaluation aspects in $\sim$35\% of cases, enabling dynamic instantiation of specialized reviewer agents, and (ii) surfaces novel evaluation dimensions absent from the fixed criteria in $\sim$30\% of cases, expanding the review space beyond predefined templates.

\begin{figure}[t]
\centering
\includegraphics[width=0.47\textwidth]{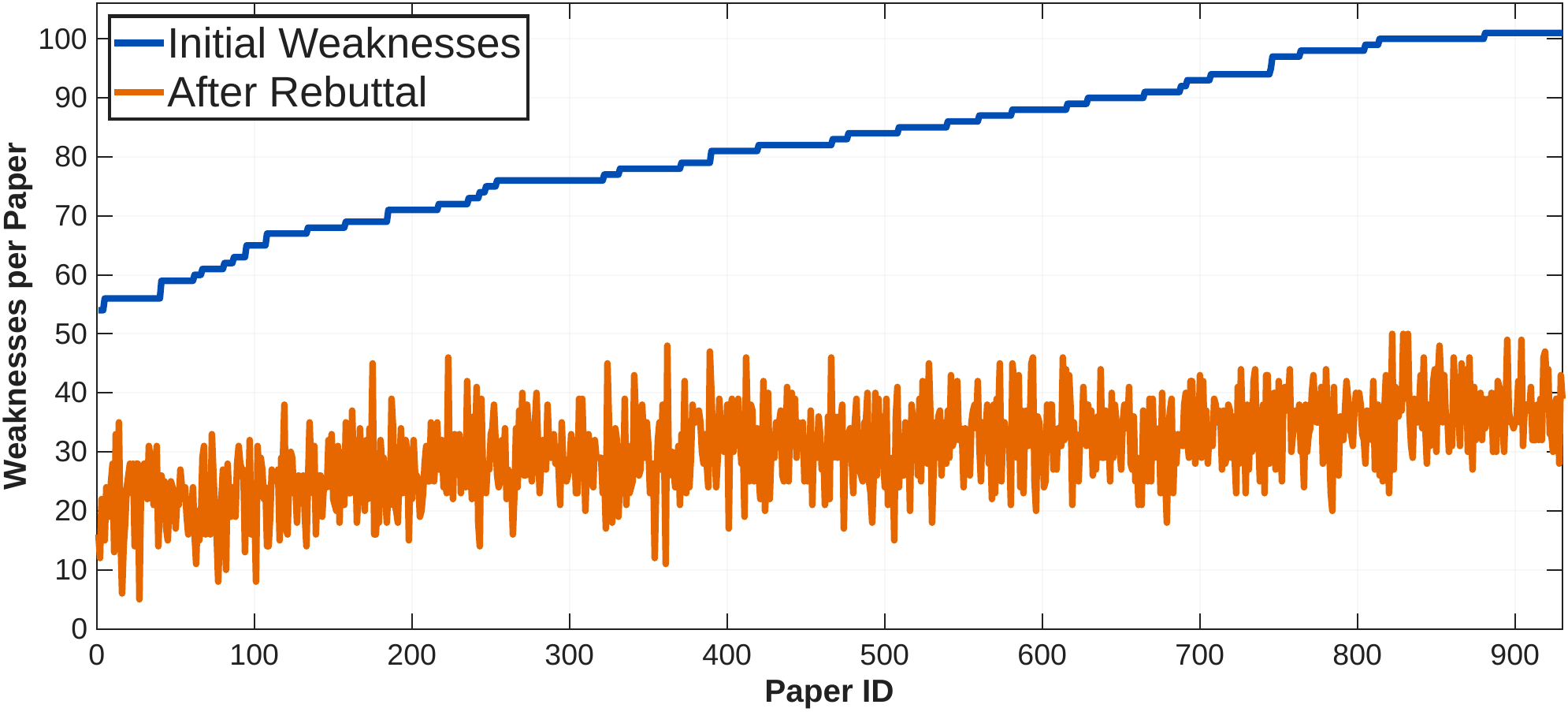} 
\vspace{-2mm}
\caption{Statistics of \#weaknesses before (blue) and after (orange) rebuttal among all test submissions.}
\label{fig:rebuttleeffect}
\vspace{-2mm}
\end{figure}
\begin{figure}[t]
\centering
\includegraphics[width=0.47\textwidth]{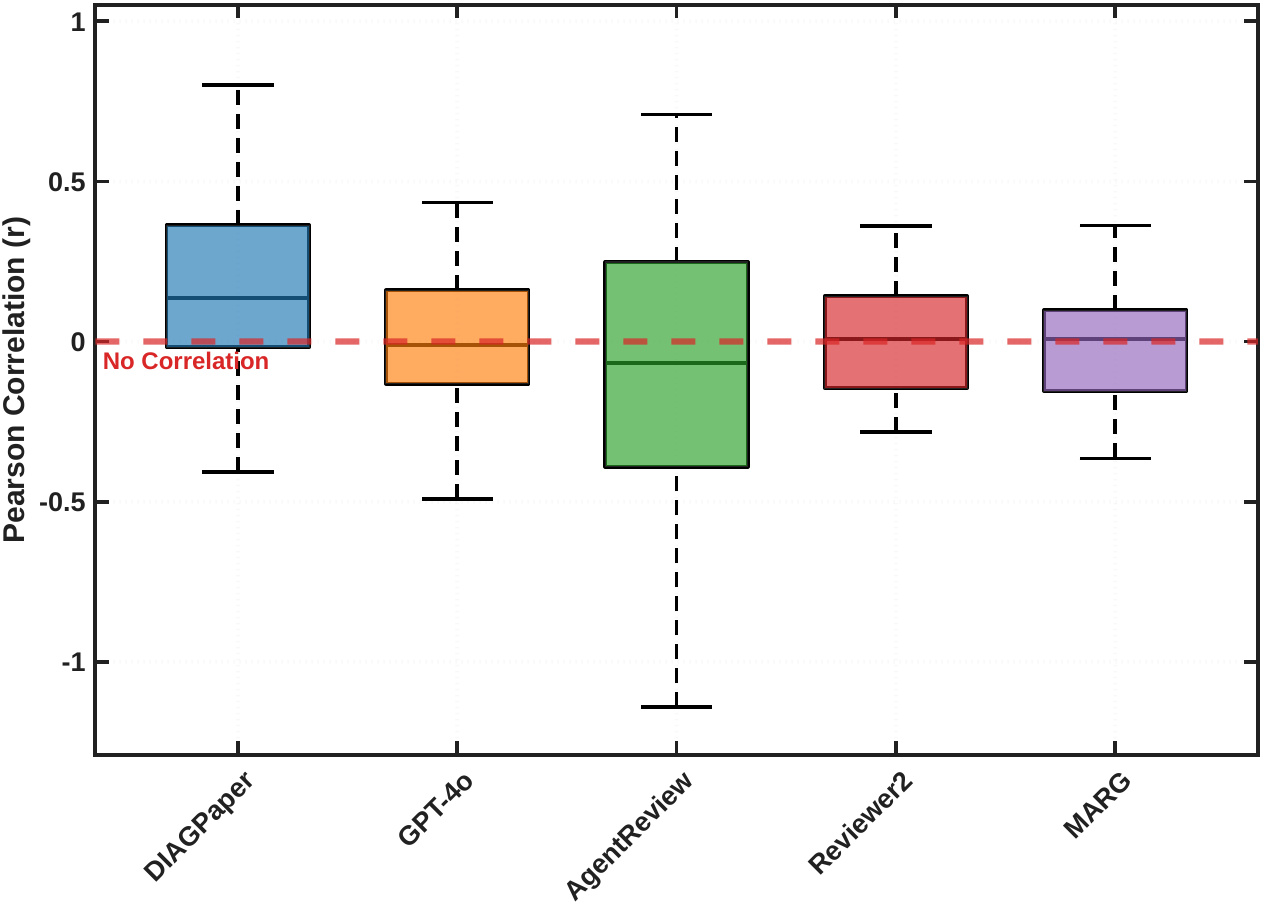} 
\vspace{-2mm}
\caption{Pearson Correlation between severity scores and F1 scores.}
\label{fig:pearson}
\vspace{-5mm}
\end{figure}

\textbf{Second, the \rebuttal{} mechanism.}
We examine whether the reviewer--author rebuttal effectively filters weak or unjustified critiques.
Using \texttt{On Pre-trained Language Models for Antibody}~\citep{wang2023pretrainedlanguagemodelsantibody} as a case study, an initial reviewer critique regarding missing alternative pre-training objectives was challenged by the rebuttal agent based on the paper’s explicit justification, experimental validation, and a clarification that the critique conflates non-exhaustiveness with poor design.
This critique received a combined validity score of 0.32, below the filtering threshold of 0.4, and was thus discarded.
For this paper, 36 initial weaknesses generated by 12 reviewer roles were reduced to 14 high-confidence ones.
Across all submissions, Figure~\ref{fig:rebuttleeffect} shows a consistent reduction from initial (blue) to retained weaknesses (orange), with an average retention rate of 40.42\%, demonstrating the intended validity-controlling behavior.


\textbf{Finally, the \prioritizer{} module.}
We evaluate whether \prioritizer{} assigns meaningful priorities to rank weaknesses.
For the same case-study paper above, prioritized weaknesses focus on methodology (35.7\%) and experiments (28.6\%), with less emphasis on reproducibility (14.3\%), aligning with the paper’s primary gaps.
To quantitatively assess prioritization, we compute the Pearson correlation between weakness rank and semantic alignment (F1) with human-written weaknesses.
As shown in Figure~\ref{fig:pearson}, \modelname{} exhibits a substantially stronger positive correlation, indicating that more severe, human-aligned weaknesses are consistently ranked earlier, whereas baseline systems show near-random rank–F1 relationships.

\begin{table}[t]
\centering
\begin{tabular}{@{}lccccc@{}}
\toprule
Method &  F1  & Sp.\\
\midrule
\modelname &50.23 & 10.25 \\ 
\enspace\enspace w/o Author rebuttal  &45.19& 9.87\\
\enspace\enspace w/o \customizer &48.63& 9.72\\ 
\enspace\enspace w/o \prioritizer & 48.46& 9.33 \\
\enspace\enspace w/o $\mathrm{Imp}^c$ in Eq. \ref{eq:damagescore}  &49.88& 10.79\\
\bottomrule
\end{tabular}
\vspace{-2mm}
\caption{Ablation Study on ReviewCritique(Valid).}
\label{tab:ablation}
\vspace{-5mm}
\end{table}

\vspace{-2mm}
\paragraph{$\mathcal{Q}_4$: Ablation Study.}
We conduct an ablation study on \textit{ReviewCritique (Valid)} to quantify the contribution of each component in \modelname{}. Specifically, we evaluate the following variants: (i)  \textit{w/o \customizer{}}, which replaces the AI-generated evaluation criteria with the 20 human-written criteria listed in Table~\ref{append:review_criteria}; (ii) \textit{w/o Author Rebuttal}, which removes the author agent from the \rebuttal{} module and retains only reviewer agents; (iii) \textit{w/o \prioritizer{}}, which disables the weakness ranking module; and (iv) \textit{w/o $\mathrm{Imp}^c$ in Eq. \ref{eq:damagescore}}

Table~\ref{tab:ablation} reports the ablation results. We observe that all components contribute positively to overall performance. Among them, removing the author rebuttal mechanism leads to the largest performance drop, indicating that explicit reviewer–author interaction is critical for filtering invalid or weakly supported critiques. Results from \textit{w/o \customizer{}} further suggest that while human-written evaluation criteria provide a strong foundation for paper review, they can be effectively replaced—and in fact improved—by AI-generated criteria that enable automated, adaptive assessment.

\section{Conclusion}
We proposed \modelname{}, a human-grounded, validity-oriented system for paper weakness identification.
By modeling key human review mechanisms, such as criterion planning, reviewer–author rebuttal, and user-oriented prioritization, \modelname{} achieves state-of-the-art performance across benchmarks.

\newpage
\section*{Limitations}
Despite its strengths, \modelname{} has several limitations. 
\begin{itemize}
    \item As a multi-agent framework, it incurs higher runtime overhead, which may restrict its applicability for large-scale submission screening.
    \item In addition, the system does not explicitly retrieve external literature, as it primarily focuses on assessing the internal consistency and self-contained validity of a paper’s claims. 
    \item  Finally, our evaluation is limited to AI-related submissions, and the effectiveness of \modelname{} on non-AI research domains remains to be explored.
\end{itemize}

\bibliography{latex/custom}

\appendix

\section{Expert-written Review Dimensions}
\label{sec:append:dimensions}

The 20 expert-written review dimensions are listed in Table \ref{append:review_criteria}.

\begin{table*}[t]
\centering
\small
\begin{tabular}{lp{12cm}}
\toprule
\textbf{Category} & \textbf{Review Dimension} \\
\midrule
\multirow{2}{*}{Importance} & \textbullet\enspace Is the problem studied in this work important? \\
& \textbullet\enspace Has the authors fully show the problem importance and is that convincing? \\
\midrule
\multirow{3}{*}{Related Work} & \textbullet\enspace Any missing related works that need to be cited and discussed? \\
 &  \textbullet\enspace If there is a Related Work section, is this section well organized? \\
 & \textbullet\enspace Has it indicated the novelties of this work compared to those prior works? \\
\midrule
\multirow{3}{*}{Clarity} &  \textbullet\enspace Do those figures, tables, etc., well support the authors' claims and align with the textual description of the authors? \\
 & \textbullet\enspace Any conflicting descriptions in this paper? \\
\midrule
Method-Novelty & \textbullet\enspace If the author claimed the approach is a newly proposed one, is the proposed approach novel? \\
Method-Clarity& \textbullet\enspace Any unclear or confusing part in the approach description? \\
\multirow{2}{*}{Method-Limitation} &\textbullet\enspace If the author claimed the approach is a newly proposed one, any limitations exist in the proposed approach but the authors didn't discuss it? \\
\multirow{2}{*}{Method-Validity} & \textbullet\enspace Any methodological flaws or inconsistencies in the proposed approach that could invalidate the results? \\
\midrule
\multirow{2}{*}{Dataset-Necessity} & \textbullet\enspace If the authors present a newly constructed dataset, does the new dataset have interesting and convincing necessities to be constructed for the target problem? \\
\multirow{4}{*}{Construction-Process} & \textbullet\enspace If the authors present a newly constructed dataset, is the construction process clear and professional? \\
 & \textbullet\enspace If the authors present a newly constructed dataset, anything should be paid special attention in the construction process and have the authors deal with them well? \\
 \multirow{2}{*}{Dataset-Representative}& \textbullet\enspace Are those datasets representative enough for this target problem? If not, any prior datasets should be tested on? \\
\midrule
\multirow{2}{*}{Exper-completeness} & \textbullet\enspace To demonstrate the effectiveness of this new approach, consider what experiments are necessary and whether the authors have conducted all the required experiments thoroughly. \\
 \multirow{2}{*}{Baseline-representative}& \textbullet\enspace Are those baselines representative enough for this target problem? If not, any missing baselines should be added, compared and discussed? \\
 \multirow{2}{*}{In-depth analysis}& \textbullet\enspace Have the experiment analyses provide enough insights or explanation or just some superficial phenomena description? \\
 \multirow{2}{*}{State-of-the-art}& \textbullet\enspace If the author claimed the approach is a newly proposed one, does the proposed approach show better performance than prior state-of-the-art? \\
 \multirow{2}{*}{Eval-metrics}& \textbullet\enspace Are those evaluation metrics for those evaluation tasks appropriate, or any cases cannot be well captured by the existing evaluation metrics? \\
\midrule
\multirow{3}{*}{Writing} & \textbullet\enspace Any grammar error in this paper? \\
 & \textbullet\enspace What severe writing issues are making this paper difficult to understand, and what remedies would you suggest? \\
\bottomrule
\end{tabular}
\caption{20 crowdsourced expert-written review dimensions are organized by category. Each dimension correspond a reviewer agent in the \rebuttal{} module.}
\label{append:review_criteria}
\end{table*}

\end{document}